\documentclass[journal]{IEEEtran}

\ifCLASSINFOpdf
\else
   \usepackage[dvips]{graphicx}
\fi
\usepackage{url}
\usepackage{graphicx}
\usepackage{amssymb}
\usepackage{amsmath}
\usepackage{multirow}
\usepackage{makecell}
\usepackage{tabularx}
\usepackage{xcolor}
\usepackage{booktabs}
\begin{document}

\title{FAMSeC: A Few-shot-sample-based General AI-generated Image Detection Method}


\author{
\IEEEauthorblockN{Juncong Xu, Yang Yang, Han Fang, Honggu Liu, and Weiming Zhang}

\thanks{This work was supported in part by the National Natural Science Foundation of China under Grant 62272003, in part by the Innovation Program for Quantum Science and Technology under Grant 2021ZD0302300. \textit{(Corresponding authors: Yang Yang; Han Fang.)}}
\thanks{Juncong Xu and Yang Yang are with Anhui University, Hefei 230039, China, and also with Institute of Artificial Intelligence, Hefei Comprehensive National Science Center, Hefei 230088, China (e-mail: wa22301178@stu.ahu.edu.cn; sky\_yang@ahu.edu.cn).}
\thanks{Han Fang is with National University of Singapore, Singapore 119077 (e-mail: fanghan@nus.edu.sg).}
\thanks{Honggu Liu and Weiming Zhang are with University of Science and Technology of China, Hefei 230026, China (e-mail: lhg9754@mail.ustc.edu.cn; zhangwm@ustc.edu.cn).}

}

\maketitle

\begin{abstract}

The explosive growth of generative AI has saturated the internet with AI-generated images, raising security concerns and increasing the need for reliable detection methods. The primary requirement for such detection is generalizability, typically achieved by training on numerous fake images from various models. However, practical limitations, such as closed-source models and restricted access, often result in limited training samples. Therefore, training a general detector with few-shot samples is essential for modern detection mechanisms. To address this challenge, we propose FAMSeC, a general AI-generated image detection method based on LoRA-based \textbf{F}orgery \textbf{A}wareness \textbf{M}odule and \textbf{Se}mantic feature-guided \textbf{C}ontrastive learning strategy. To effectively learn from limited samples and prevent overfitting, we developed a forgery awareness module (FAM) based on LoRA, maintaining the generalization of pre-trained features. Additionally, to cooperate with FAM, we designed a semantic feature-guided contrastive learning strategy (SeC), making the FAM focus more on the differences between real/fake image than on the features of the samples themselves. Experiments show that FAMSeC outperforms state-of-the-art method, enhancing classification accuracy by 14.55\% with just 0.56\% of the training samples.

\end{abstract}


\IEEEpeerreviewmaketitle

\section{Introduction}

\IEEEPARstart{T}{he} rapid evolution of generative models, such as generative adversarial networks \cite{cyclegan, progan, biggan} and diffusion models \cite{ddpm, ldm, glide, MirrorDiffusion}, has led to the creation of AI-generated images that exhibit remarkable realism. However, these advancements in generative AI also raise concerns regarding security and privacy in human society. Consequently, there is a growing demand for the development of detection methods capable of identifying AI-generated images.

    The diversity of generation mechanisms leads to a broad spectrum of images produced by different models, highlighting the essential requirement for any detection mechanism: generalizability. One straightforward way to ensure generalization is to train a detector with a large collection of fake images generated by various models. However, collecting data is time-consuming and laborious. In practical applications, we may only have access to a limited number of training samples due to closed-source models and access restrictions, such as those from the DALL-E~\cite{dalle} series and Midjourney~\cite{midjourney}. Therefore, achieving good generalization with the few-shot sample is the practical and fundamental requirement for modern detectors. 
    

    Most current AI-generated image detection methods rely heavily on large amounts of training data to achieve generalization~\cite{wang, spldetect1, spldetect2, lgrad}. For example, Wang~\textit{et al.}~\cite{wang} trained a classifier on a dataset of 720,000 real and fake images, using data augmentation to improve generalization. Similarly, Tan~\textit{et al.}~\cite{lgrad} used the same dataset to train a classifier based on differences in gradient information between real and fake images processed by a pre-trained model. While these methods perform well with a large number of training samples, they struggle to generalize effectively when training samples are limited. 

     To solve this problem, this paper introduces FAMSeC, a general AI-generated image detector that requires only a small number of training samples. Our method builds on the pre-trained CLIP:ViT's features, which have been proved to provide sufficient generalizable features to realize detection across models~\cite{unifd}. The biggest challenge is how to fine-tune the model to retain the general features and avoid overfitting problems while at the same time learning useful features through few-shot samples.

     
     
     To address such a challenge, we propose a forgery awareness module (FAM) based on Low-Rank Adaptation (LoRA)~\cite{lora}, which can effectively mitigate dramatic changes in pre-trained features while ensuring the sufficient learning of the few-shot samples, thus both achieving the cross-model generalizability and updating the discriminating features with few-shot samples. Besides, to guide FAM to learn more general features, we developed a semantic feature-guided contrastive learning strategy (SeC). This strategy uses the rich semantic feature extracted by a pretrained CLIP:ViT to create positive/negative sample pairs with the features extracted by another FAM-enhanced CLIP:ViT, making the FAM focus more on enlarging the difference between real and fake images rather than on learning the features of the samples themselves.
     
     
     
     Experiments demonstrate that our model, trained with only 4,000 real and fake images from the ProGAN dataset, achieved an average classification accuracy of 95.22\% across three cross-model datasets, which include a variety of unseen GAN and diffusion models. Compared to state-of-the-art method, our model uses just 0.56\% of the training samples and improves detection accuracy by 14.55\%.

     In summary, the contributions in this paper are as follows:
     \begin{itemize}
    \item[$\bullet$] To achieve robust generalization for AI-generated image detection with few samples, we designed a forgery awareness module (FAM) based on LoRA that effectively adapts CLIP:ViT for extracting discriminative features of AI-generated images while preserving the generalization of the pre-trained features.
    \item[$\bullet$] We designed a semantic feature-guided contrastive learning strategy (SeC) that cooperates with the proposed FAM to enable it to learn the general differences between real and fake images, rather than the specific features of the training samples.
    \item[$\bullet$] Experiments show that our proposed method uses only 0.56\% of the training data required by current state-of-the-art method, achieving an average detection accuracy improvement of 14.55\%.
    \end{itemize}

\section{Proposed method}


\subsection{Motivation and Overview}
\begin{figure}[h]
  \centering
  \includegraphics[width=\linewidth]{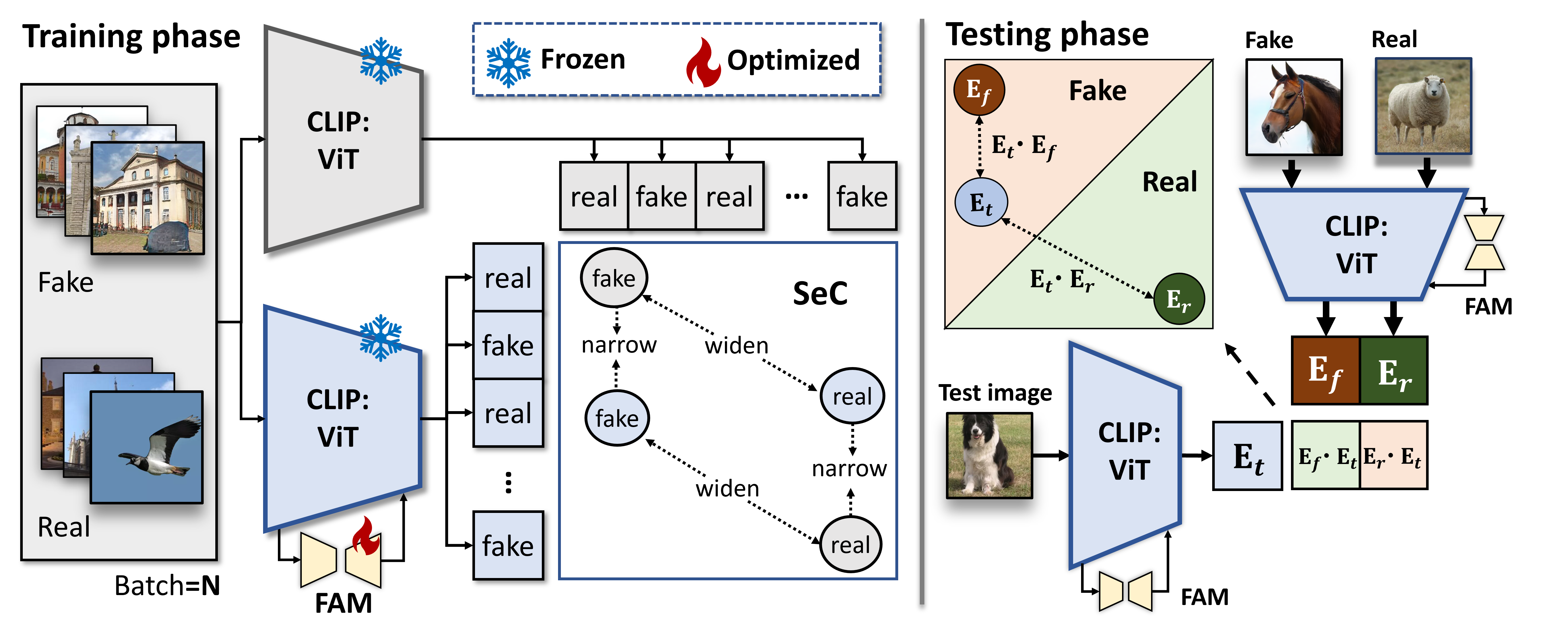}
  \caption{The framework of our proposed FAMSeC. During the training phase, we use two CLIP:ViT to perform semantic feature-guided contrastive learning (SeC). One CLIP:ViT with fixed parameters is used to extract semantically rich features to guide the contrastive learning, while the other CLIP:ViT acts as a feature extractor enhanced by a LoRA-based forgery awareness module (FAM) to learn the differences between real and fake images. During the testing phase, the features of the input image, extracted by the feature extractor, are compared with the features of real and fake images to measure the distance and derive the prediction results.}
  \label{arch}
\end{figure}

Our goal is to develop an AI-generated image detector that can achieve robust cross-model generalization in the scenario of few-shot samples. To achieve this, we used CLIP:ViT-L/14 as feature extractor and introduced a forgery awareness module (FAM) based on LoRA to prevent CLIP:ViT from overfitting to the few training samples. Additionally, to ensure FAM focuses more on the distinction between real and fake images rather than on the training samples themselves, we developed a semantic feature-guided contrastive learning strategy. The overall framework of FAMSeC is shown in Fig. \ref{arch}.



\begin{figure}[h]
  \centering
  \includegraphics[width=0.85\linewidth]{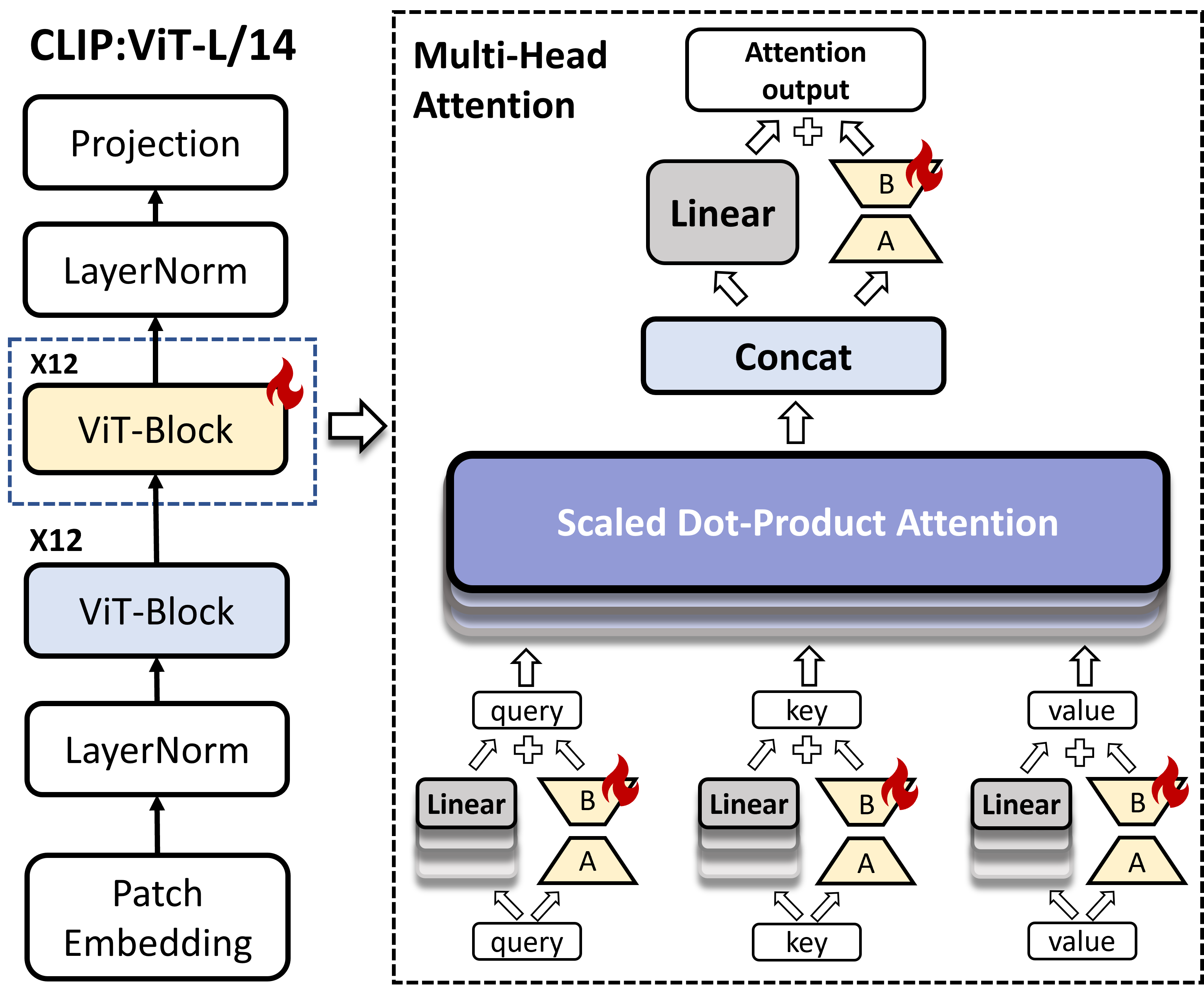}
  \caption{Diagram of the LoRA-based forgery awareness module (FAM). The LoRA is applied to the $query$, $key$, $value$, and $output$ matrices of the multi-head attention modules in the last 12 ViT blocks of CLIP:ViT-L/14.}
  \label{lora}
\end{figure}


\subsection{Forgery awareness module (FAM)}
To adapt CLIP:ViT for AI-generated image detection while preventing overfitting, we introduce a forgery awareness module (FAM) based on LoRA, as shown in Fig. \ref{lora}. We applied LoRA~\cite{lora} to the multi-head attention modules of the last 12 ViT blocks in CLIP:ViT to enhance the model's awareness of the differences between real and fake images. In each multi-head attention module, the $query$, $key$, $value$, and $output$ matrices are modified using the LoRA modules with a rank of 2. Specifically, let $W_0 \in \mathbb{R}^{d \times k}$ represent any of the pretrained matrices mentioned above, we constrain its update by representing it with a low-rank decomposition:
\begin{equation}
W_0 + \Delta W = W_0 + BA,
\end{equation}
where $B \in \mathbb{R}^{d \times r}$, $A \in \mathbb{R}^{r \times k}$ and the rank $r \ll \min(d, k)$. During the training process, the parameters of $W_0$ are frozen and do not participate in updates, while the parameters of matrices $A$ and $B$ are trainable.

\subsection{Semantic feature-guided contrastive learning (SeC)}
To guide the FAM in learning more general features, we designed SeC to make FAM focus more on the differences between real and fake images rather than on the samples themselves, as is shown in Fig. \ref{arch}. The detailed procedure can be described as: We first collect the well-labeled training data $\mathbb{X}$ with corresponding label $\mathbb{Y}$, which contains the real images $x_{real}$ with label $y_{real}$, and the fake images $x_{fake}$ with label $y_{fake}$. $\{x_{real},x_{fake}\} \in \mathbb{X}, \{y_{real},y_{fake}\} \in \mathbb{Y}$, where $x_{fake}$ are generated with specific generation models. In this paper, $y_{real} = \textbf{1}, y_{fake} = \textbf{0}$. Then we employ two pretrained CLIP:ViT for training. One of them is used as the guiding model, denoted as $G$, and the other one, $T$, serves as the feature extractor and is enhanced with LoRA-based forgery awareness modules. In each training batch, $N$ images ($N$ indicates the training batch size) along with other labels $\{ x_i; y_i \mid 1 \leq i \leq N\}\in\{\mathbb{X};\mathbb{Y}\}$ are selected. Then every image in the batch $x_i$ is fed into $G$ and $T$ respectively to obtain the respective embeddings $E^G_i = G(x_i)$ and $E^T_i = T(x_i)$.

Subsequently, we generate $N\times N$ feature pairs based on $E^G_i$ and $E^T_j$, where $1 \leq i \leq N, 1 \leq j \leq N$. Then we assign a similarity score $p_{i,j}$ to each pair, where $p_{i,j}$ can be calculated by:
\begin{equation}
p_{i,j} = \frac{E^G_i \cdot E^T_j }{\Vert E^G_i \Vert \Vert E^T_j\Vert },
\end{equation}
where $\circ$ indicates the dot product. We assign a label $l_{i,j}$ to each $p_{i,j}$ with the following manner:

\begin{equation}
l_{i,j} = y_i \odot y_j,
\end{equation}
where $\odot$ indicates the exclusive XNOR operation. Then the combination of $p_{i,j}$ and $l_{i,j}$ are utilized to update low-rank matrices within the FAM according to the loss functions:

\begin{equation}
\begin{split}
\mathcal{L}_{\text{con}} = & - \frac{1}{N \cdot N} \sum_{i=1}^{N} \sum_{j=1}^{N} \left[ l_{i,j} \cdot \log\left(\sigma\left(\frac{p_{i,j}}{\tau}\right)\right) \right. \\
& \left. + (1 - l_{i,j}) \cdot \log\left(1 - \sigma\left(\frac{p_{i,j}}{\tau}\right)\right) \right],
\end{split}
\end{equation}
where $\tau$ represents a learnable temperature coefficient.

\subsection{Testing}
In the testing phase, we can determine whether a test image is real or fake by analyzing the distance between its embeddings and those of real and fake images from training set. As shown in Fig. \ref{arch}, the test image’s embedding, denoted as $E_t$,  is extracted by the feature extractor. The cosine distances $d_f$ and $d_r$ between $E_t$ and the embeddings of real and fake images ($E_f$ and $E_r$) can be calculated as follows:
\begin{equation}
d_f = \frac{E_f \cdot E_t }{\Vert E_f \Vert \Vert E_t\Vert },  d_r = \frac{E_r \cdot E_t }{\Vert E_r \Vert \Vert E_t\Vert }.
\end{equation}

The model's prediction can be determined based on the relationship between $d_r$ and $d_r$:
\begin{equation}
\hat{l} =
\begin{cases}
fake & \text{if } d_f > d_r \\
real & \text{if } d_f \leq d_r
\end{cases}.
\end{equation}


\section{Experiment}

\subsection{Datasets and Implementation Details}
\textbf{Datasets }
The training set is from the ForenSynths dataset provided by Wang \textit{et al.}~\cite{wang}, which includes 720k images consisting of fake images generated by ProGAN~\cite{progan} and real images from LSUN~\cite{lsun}. We randomly selected a subset of samples (e.g., 4,000 images) from the training set to train FAMSeC. We used three cross-model datasets for testing: ForenSynths~\cite{wang}, UniversalFakeDetect~\cite{unifd}, and GenImage~\cite{genimage} datasets. The ForenSynths dataset includes seven types of GAN datasets: ProGAN~\cite{progan}, CycleGAN~\cite{cyclegan}, BigGAN~\cite{biggan}, StyleGAN~\cite{stylegan}, StyleGAN2~\cite{stylegan2}, GauGAN~\cite{gaugan}, and StarGAN~\cite{stargan}. The UniversalFakeDetect dataset contains three types of diffusion models: Guided~\cite{guided}, LDM~\cite{ldm}, and Glide~\cite{glide}, and one autoregressive model: DALL-E~\cite{dalle}. The GenImage dataset includes six types of diffusion models: Midjourney (MJ)~\cite{midjourney}, Stable Diffusion (SD)~\cite{ldm}, ADM~\cite{guided}, Glide~\cite{glide}, Wukong~\cite{wukong}, and VQDM~\cite{vqdm}. 

\textbf{Implementation Details }We employed Adam as the optimizer, using a learning rate of 1e-4 to optimize the parameters of the FAM. In FAM, each LoRA module has a rank of 2, and the dropout probability is set to 0.25. All input images are randomly cropped to a size of 224 $\times$ 224. During the testing phase, the real/fake images used for distance-based classification are randomly selected from the training set.

The experiments were conducted on a server equipped with Intel Xeon Gold 6230 (2.10 GHz) $\times$ 2 and NVIDIA RTX A6000 $\times$ 4, with a total running memory of 512GB.

\begin{table*}
  \caption{The results (Accuracy) of our method compared to the baselines on three cross-model datasets.}
  \label{ACC}
    \belowrulesep=0pt
    \aboverulesep=0pt
  \setlength{\tabcolsep}{2.1pt}
  \fontsize{6pt}{8pt}\selectfont
  \begin{tabular}{ccccccccccccccccccccccccc} 
    \toprule
\multirow{2}{*}{\makecell{Detection\\method}} & \multirow{2}{*}{\makecell{Training\\samples}} & \multicolumn{7}{c}{ForenSynths~\cite{wang}} & \multicolumn{8}{c}{UniversalFakeDetect~\cite{unifd}} & \multicolumn{7}{c}{GenImage~\cite{genimage}} & \multirow{2}{*}{\makecell{Avg.\\ACC}} \\
    \cmidrule(lr){3-9} \cmidrule(lr){10-17} \cmidrule(lr){18-24}
    & & \makecell{Pro-\\GAN} & \makecell{Cycle-\\GAN} & \makecell{Big-\\GAN} & \makecell{Style-\\GAN} & \makecell{Style-\\GAN2} & \makecell{Gau-\\GAN} & \makecell{Star-\\GAN} & Guided & \makecell{LDM-\\200} & \makecell{LDM\\200/CFG} & \makecell{LDM\\ 100} & \makecell{Glide\\100/27} & \makecell{Glide\\50/27} & \makecell{Glide\\100/10} & DALL-E & MJ & SD1.4 & SD1.5 & ADM & GLIDE & Wukong & VQDM & \\
    \midrule
    CNNDet & 720k & \textbf{100.0 }& 85.18 & 70.45 & 85.58 & 82.88 & 78.09 & 92.11 & 63.65 & 53.85 & 55.20 & 55.10 & 60.30 & 62.70 & 61.00 & 56.05 & 53.06 & 51.51 & 51.26 & 58.91 & 56.34 & 49.22 & 54.91 & 65.33\\ 
    \midrule
    No-down & 720k &\textbf{100.0} & 88.67 & 89.66 & 93.19 & 90.65 & 90.53 & 90.53 & 61.57 & 56.49 & 58.97 & 56.74 & 63.05 & 67.63 & 64.66 & 62.12 & 53.71 & 54.88 & 56.49 & 52.41 & 60.14 & 52.16 & 69.18 & 69.70 \\ 
    \midrule
    LNP & 720k & 99.78 & 83.40 & 81.83 & 91.39 & 93.16 & 70.25 & \textbf{99.88} & 66.85 & 79.70 & 81.30 & 80.60 & 76.50 & 77.90 & 80.10 & 83.30 & 62.62 & 80.25 & 79.42 & 78.34 & 78.31 & 77.50 & 67.97 & 80.47\\ 
    \midrule
    LGrad & 720k & 99.81 & 85.53 & 82.05 & 89.69 & 86.23 & 80.84 & 98.08 & 81.05 & 87.05 & 88.25 & 88.25 & 87.30 & 90.35 & 90.70 & 86.20 & \textbf{67.35} & 63.02 & 64.17 & 61.44 & 70.76 & 58.57 & 67.82 & 80.66\\ 
    \midrule
    DIRE &720k& \textbf{100.0} & 66.53 & 67.12 & 84.32 & 75.63 & 66.47 & 98.69 & 84.25 & 83.47 & 84.66 & 84.21 & 88.01 & 91.32 & 90.46 & 59.21 & 58.35 & 49.63 & 49.76 & 76.36 & 72.41 & 55.39 & 54.37 & 74.57\\
    \midrule
    UniFD &720k& 99.87 & \textbf{98.46} & \textbf{95.28} & 85.96 & 73.10 & \textbf{99.50} & 96.62 & 70.40& 94.17 & 73.59 & 95.34 & 78.37 & 78.24 & 78.85 & 86.94 & 57.35 & 61.15 & 62.99 & 68.07 & 64.01 & 71.32 & 85.17 & 80.67\\ 
    \midrule
    Ours &4k& \textbf{100.0} & 95.65 & 94.18 & \textbf{99.00} & \textbf{98.10} & 90.01 & 98.71 & \textbf{88.91} & \textbf{99.45} & \textbf{99.26} & \textbf{99.33} & \textbf{98.35} & \textbf{99.14} & \textbf{99.02} & \textbf{99.33} & 65.07 & \textbf{97.98} & \textbf{97.18} & \textbf{87.12} & \textbf{96.32} & \textbf{97.06} & \textbf{95.59} & \textbf{95.22} \\
  \bottomrule
\end{tabular}
\end{table*}

\subsection{Cross-model Experiment}
In this experiment, our model will be compared with six baseline methods: CNNDet (CVPR'2020)~\cite{wang}, No-down (ICME'2021)~\cite{nodown}, LNP (ECCV'2022)~\cite{liu}, LGrad (CVPR'2023)~\cite{lgrad}, DIRE (ICCV'2023)~\cite{dire}, and UniFD (CVPR'2023)~\cite{unifd}. All models are trained using the ProGAN dataset from ForenSynths~\cite{wang} dataset and tested on three cross-model datasets. 
Note that all baseline models were trained using the complete training set, while our FAMSeC was trained using only 4,000 training samples.








Table \ref{ACC} presents the accuracy (ACC) of FAMSeC and baseline methods on three cross-model datasets. It can be observed that our FAMSeC achieves the highest classification accuracy in the ForenSynths, UniversalFakeDetect, and GenImage. The average classification accuracy across these three datasets is 14.55\% higher than that of the best-performing baseline method, UniFD. In addition, we observed that all methods demonstrated high accuracy on the ForenSynths dataset. However, in the UniversalFakeDetect and GenImage datasets, most baseline methods experienced some degree of decline in performance. Particularly in the GenImage dataset, the best-performing baseline method in this dataset, LNP, achieved only 74.92\% classification accuracy. In contrast, our FAMSeC exhibited consistently high ACC across these datasets, with average accuracies of 96.52\%, 97.85\%, and 90.90\%, respectively, showing better generalization capabilities.


\subsection{Ablation Study}
\subsubsection{Effectiveness of components}

\begin{table}[tb]
  \caption{Ablation results (Accuracy) of the LoRA-based forgery awareness module (FAM) and semantic feature-guided contrastive learning (SeC).}
  \label{ablation}
  \setlength{\tabcolsep}{4.0pt}
  \centering
    \belowrulesep=0pt
    \aboverulesep=0pt
  \begin{tabular}{cc|cccc} 
    \toprule
    \makecell{FAM} & SeC & \makecell{Foren-\\Synths~\cite{wang}} & \makecell{Universal-\\FakeDetect~\cite{unifd}} & GenImage~\cite{genimage} & Avg.\\
    \midrule
    $\times$ & $\times$ & 92.09 & 93.29 & 74.26 & 86.55 \\
    \midrule
    \checkmark & $\times$ & \textbf{97.60} & 91.58 & 77.71 & 88.96\\
    \midrule
     $\times$ & \checkmark& 88.98 & 97.06 & 86.34 & 90.79\\
    \midrule
    \checkmark  & \checkmark &96.52 & \textbf{97.85} & \textbf{90.90} & \textbf{95.09}\\
    \bottomrule
  \end{tabular}
\end{table}

We conducted ablation experiments to evaluate the effectiveness of the FAM  and the SeC. The results are shown in Table \ref{ablation}. The first row of the table represents the model fully fine-tuned using classification loss on CLIP:ViT. The data indicates that introducing either the FAM or the SeC alone improves the average accuracy by 2.41\% and 4.24\%, respectively, compared to the fully fine-tuned model. Our proposed FAMSeC, which combines both FAM and SeC, achieved the highest classification accuracy, outperforming the fully fine-tuned model by 8.54\%. This validates the effectiveness of our approach.

\begin{table}[h]
  \caption{Ablation results (Accuracy) of the applied ViT blocks of FAM.}
  \label{apply_blocks}
      \belowrulesep=0pt
    \aboverulesep=0pt
  \centering
  \begin{tabular}{c|cccc} 
    \toprule
    \makecell{Adapted\\blocks} &  \makecell{Foren-\\Synths~\cite{wang}} & \makecell{Universal-\\FakeDetect~\cite{unifd}} & GenImage~\cite{genimage} & Avg.\\
    \midrule
    Last 6  & 95.96 & 87.94 & 75.03 & 86.31\\
    \midrule
    Last 12  & \textbf{96.52} & 97.85 & \textbf{90.90} & \textbf{95.09}\\
    \midrule
    Last 18  & 94.42 & \textbf{98.09} & 89.07 & 93.86\\
    \midrule
    Last 24  & 95.37 & 97.75 & 89.57 &94.23\\
    \bottomrule
  \end{tabular}
\end{table}

\subsubsection{Impact of the application range of the FAM}
As shown in Table \ref{apply_blocks}, applying more ViT blocks with FAM is not necessarily better. The model achieved the highest average accuracy when FAM was applied to the last 12 ViT blocks of CLIP:ViT. When FAM was applied to more ViT blocks, accuracy slightly decreases. This decline is due to the increased learning capacity, which makes the model more prone to overfitting the training set. Additionally, when FAM was applied to only the last 6 ViT blocks, the model's average accuracy was just 86.31\%. This is because FAM's learning capacity and its impact on the pretrained weights are too small to effectively capture the differences between real/fake images.

\begin{table}[h]
  \caption{Ablation results (Accuracy) of the LoRA rank.}
  \label{rank}
      \belowrulesep=0pt
    \aboverulesep=0pt
  \centering
  \begin{tabular}{c|cccc} 
    \toprule
    \makecell{LoRA\\rank} & ForenSynths~\cite{wang} & \makecell{Universal-\\FakeDetect~\cite{unifd}} & GenImage~\cite{genimage} & Avg.\\
    \midrule
    2 & 96.52 & \textbf{97.85} & \textbf{90.90} & \textbf{95.09}\\
    \midrule
    4  & 97.92 & 95.86 & 89.04& 94.27\\
    \midrule
    8  & \textbf{98.70} & 95.99 & 87.34& 94.01\\
    \midrule
    16  & 98.05 & 95.58 & 88.06 & 93.90\\

    \bottomrule
  \end{tabular}
\end{table}

\subsubsection{Impact of the rank of LoRA}
As shown in Table \ref{rank}, the model achieved the best performance when the rank of LoRA was set to 2. It can be observed that as the rank gradually increases, the model's average ACC declines. This is due to an increase in the number of learnable parameters as the rank increases, making the model more prone to overfitting on the training set, thereby reducing generalizability.

 \begin{figure}[tb]
  \centering
  \includegraphics[width=0.8\linewidth]{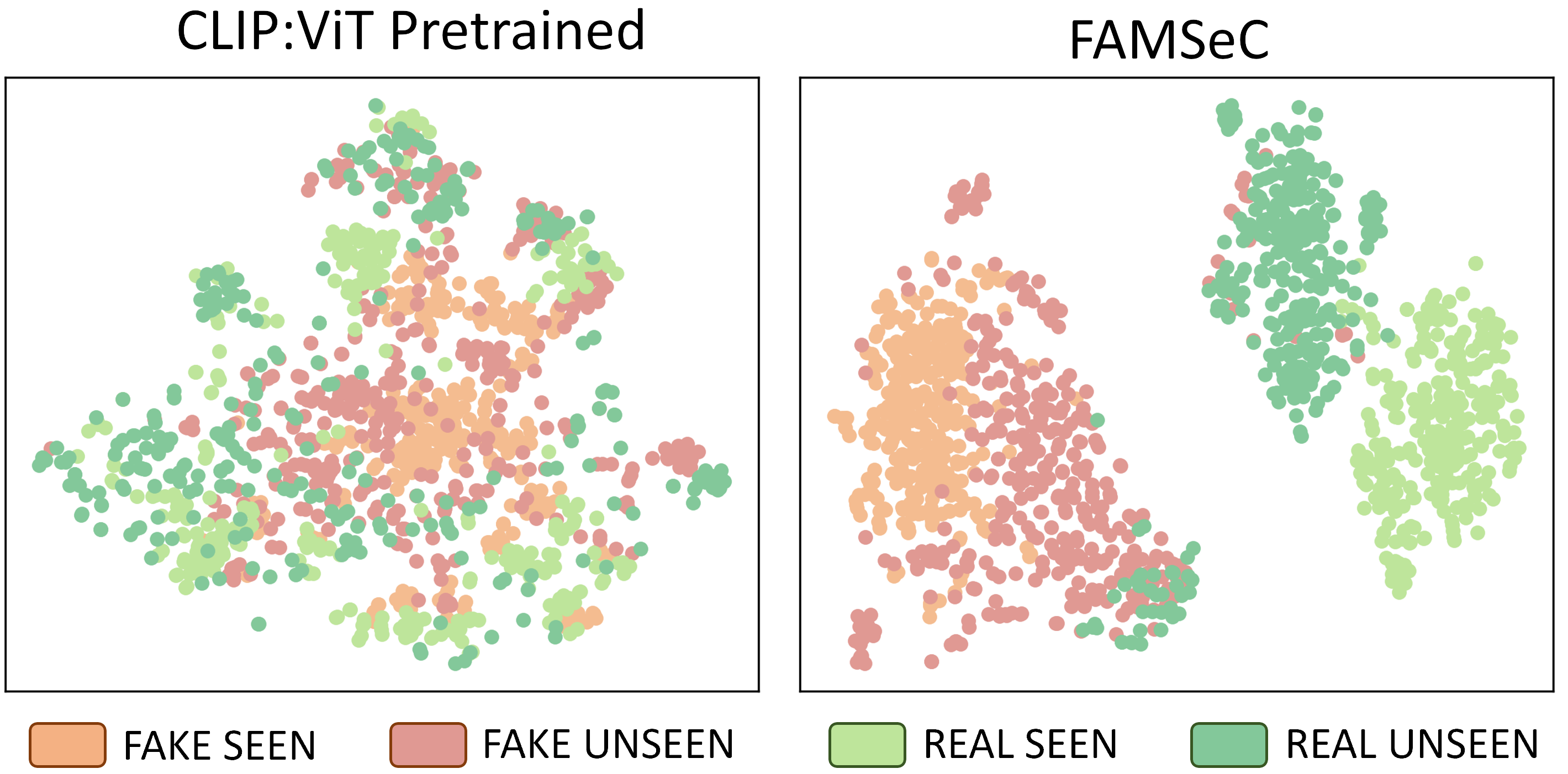}
  \caption{The t-SNE visualization of the feature space for the pretrained CLIP:ViT-L/14 and our FAMSeC.}
\label{tsne}
\end{figure}

 \begin{figure}[h]
  \centering
  \includegraphics[width= \linewidth]{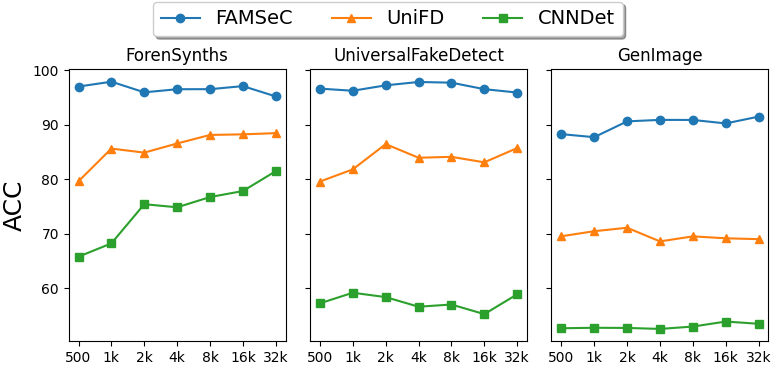}
  \caption{The detection accuracy of our FAMSeC, UniFD~\cite{unifd}, and CNNDet~\cite{wang} across three cross-model datasets with different training sample sizes. Note that all models are trained using the training set from the ForenSynthst~\cite{wang} dataset.}
 \label{datalen}
\end{figure}

\subsection{Visualization Results}
To further verify the effectiveness of FAMSeC, we visualized its feature space using t-SNE~\cite{tsne}, as shown in Fig. \ref{tsne}. In the feature space of the pretrained CLIP:ViT, the features of real and fake images are intertwined, making it difficult to find an effective classification boundary. In contrast, in the feature space of our proposed FAMSeC, real and fake image features exhibit distinct distributions, with a significant margin for classification between them. Moreover, features of unseen real and fake images align closely with their respective visible image categories, demonstrating our method's robust generalization to unseen generative models.

\subsection{The Impact of Training Sample Size}

Fig. \ref{datalen} displays the detection performance of FAMSeC, UniFD~\cite{unifd}, and CNNDet~\cite{wang} across three datasets with different training sample sizes. Our FAMSeC consistently outperforms UniFD~\cite{unifd} and CNNDet~\cite{wang} in ACC across various training sample sizes, particularly on the more challenging UniversalFakeDetect and GenImage datasets.

\section{Conclusion}
In this paper, we propose FAMSeC, a general AI-generated image detection method that can achieve strong generalization with only a small number of training samples. Based on CLIP:ViT, FAMSeC learns to distinguish the general and essential difference between real and fake images by using a LoRA-based forgery awareness module and a semantic feature-guided contrastive learning strategy. Experiments show that our FAMSeC achieves impressive cross-model detection performance by training on only a limited number of samples from the ProGAN dataset. This capability extends not only to various unseen GAN models but also to various unseen models in the diffusion family, demonstrating the effectiveness of our proposed method.



\bibliographystyle{IEEEtran}
\bibliography{IEEEabrv,sample-base}

\end{document}